\documentclass[journal]{IEEEtran}
\usepackage{graphicx}
\usepackage{subfigure}
\usepackage{multirow}
\usepackage{color}
\usepackage{times}
\usepackage{soul}
\usepackage{url}
\usepackage{amsthm}
\usepackage{booktabs}
\usepackage{algorithm}
\usepackage{algorithmic}
\usepackage{threeparttable}
\usepackage{ulem}
\usepackage[switch]{lineno}
\usepackage{amsfonts}
\usepackage{amsmath}
\usepackage{mathtools}
\usepackage{cancel} 
\usepackage{caption}
\usepackage{hyperref}
\hypersetup{
	colorlinks=true,
	linkcolor=blue,
	filecolor=blue,      
	urlcolor=blue,
	citecolor=cyan,
}

\usepackage{array}
\usepackage{colortbl}

{}
{}
{}

\ifCLASSINFOpdf
\else
\fi
\begin{document}
		%
		\title{Multi-direction and Multi-scale Pyramid in Transformer for Video-based Pedestrian Retrieval}
		%
		%
		%
		
		\author{Xianghao~Zang,
			Ge~Li,
			and~Wei~Gao
			\thanks{This work was supported by the National Key R\&D Program of China (2020AAA0103501). (\textit{Corresponding author: Wei Gao.})}
			\thanks{Xianghao Zang, Ge Li and Wei Gao are with School of Electronic and Computer Engineering, Peking University, Shenzhen 518055, China (e-mail: zangxh@pku.edu.cn; geli@ece.pku.edu.cn; gaowei262@pku.edu.cn).}
		}
	
	%
	%

\markboth{Journal of \LaTeX\ Class Files,~Vol.~14, No.~8, August~2015}%
{Shell \MakeLowercase{\textit{et al.}}: Bare Demo of IEEEtran.cls for IEEE Journals}
%

\IEEEpubid{\begin{minipage}{\textwidth}\ \\ \\ \\ \\
	0000--0000/00\$00.00~\copyright~2022 IEEE. Personal use of this material is permitted. Permission from IEEE must be obtained for all other uses, in any current or future media, including reprinting/republishing this material or promotional purposes, creating new collective works, for resale or redistribution to servers or lists, or reuse of any copyrighted component of this work in other works.
\end{minipage}}


\maketitle
\begin{abstract}
	In video surveillance, pedestrian retrieval (also called person re-identification) is a critical task. This task aims to retrieve the pedestrian of interest from non-overlapping cameras. Recently, transformer-based models have achieved significant progress for this task. However, these models still suffer from ignoring fine-grained, part-informed information. This paper proposes a multi-direction and multi-scale Pyramid in Transformer (PiT) to solve this problem. In transformer-based architecture, each pedestrian image is split into many patches. Then, these patches are fed to transformer layers to obtain the feature representation of this image. To explore the fine-grained information, this paper proposes to apply vertical division and horizontal division on these patches to generate different-direction human parts. These parts provide more fine-grained information.  To fuse multi-scale feature representation, this paper presents a pyramid structure containing global-level information and many pieces of local-level information from different scales. The feature pyramids of all the pedestrian images from the same video are fused to form the final multi-direction and multi-scale feature representation. Experimental results on two challenging video-based benchmarks, MARS and iLIDS-VID, show the proposed PiT achieves state-of-the-art performance. Extensive ablation studies demonstrate the superiority of the proposed pyramid structure. The code is available at \href{https://git.openi.org.cn/zangxh/PiT.git}{https://git.openi.org.cn/zangxh/PiT.git}.
	
\end{abstract}


%
\IEEEpeerreviewmaketitle

\section{Introduction}
%
%
%
%
\IEEEPARstart{P}{edestrian} retrieval is a critical task in intelligent surveillance \cite{RN632} \cite{RN636} \cite{RN687}. Given a pedestrian image as the query, pedestrian retrieval aims to find the right images in a large gallery. The query image and the matched gallery images must be from different cameras. Pedestrian retrieval has important practical applications in both society and industry, such as finding the criminal suspects and tracking pedestrian movement. Compared to the image-based pedestrian retrieval, the video-based one can provide much more gait and view information of pedestrians and alleviate the negative effects of occlusion situations. Therefore, video-based pedestrian retrieval is getting more and more attention from researchers \cite{RN635}.

For CNN-based pedestrian retrieval, dividing the feature map into multiple horizontal stripes and individually training each one are general operations. After the training is complete, all the stripe features are assembled to generate the convolution descriptor for each image, which provides the model a rich feature representation \cite{RN191}. Based on the horizontal division strategy, a pyramid of stripes is proposed to exploit the partial information of each pedestrian \cite{RN318}. Although these methods above improve the model performance, the direction of division strategy is limited.

\begin{figure}[]
	\centering
	\includegraphics[width=0.45\textwidth]{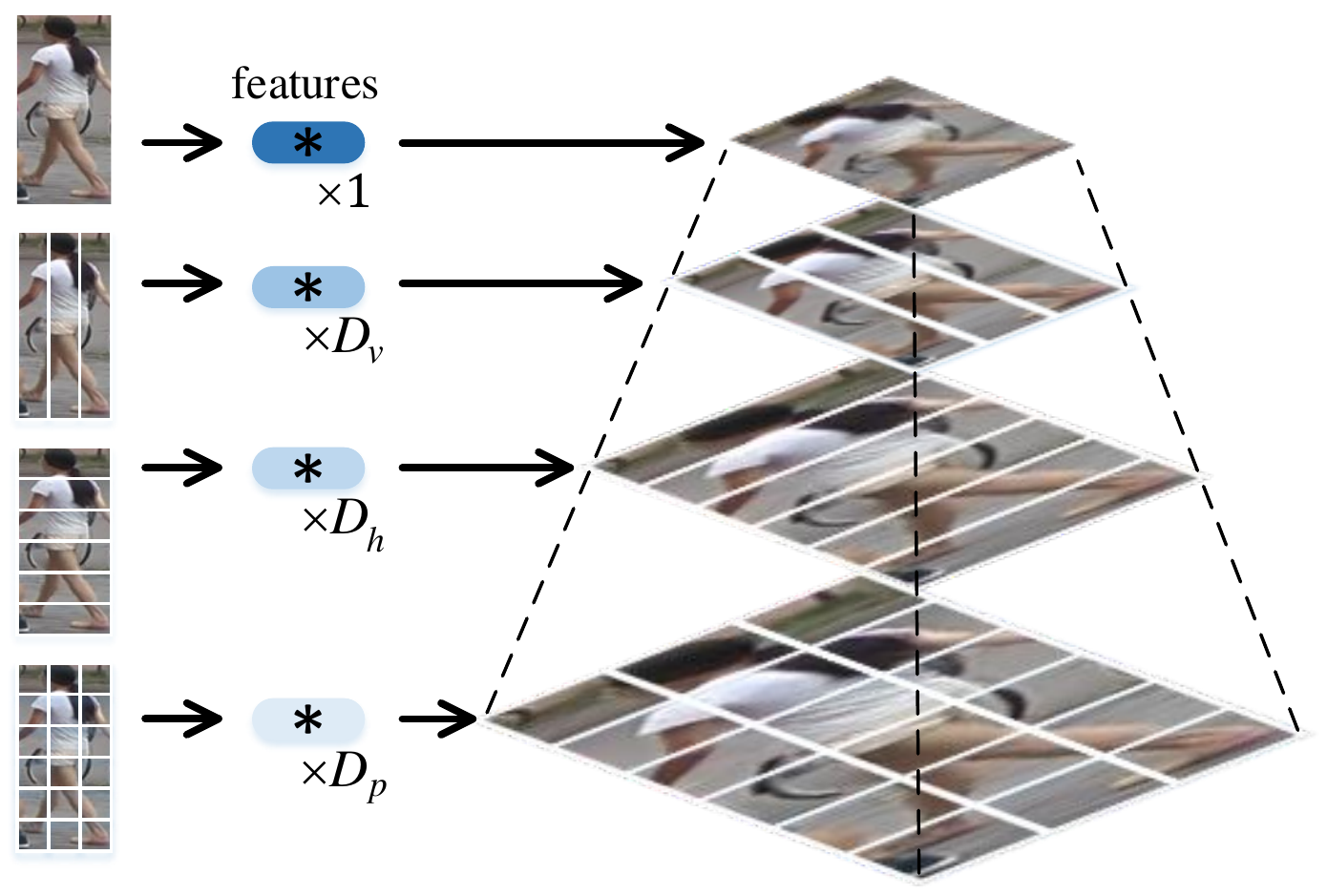}
	\caption{Multi-direction and multi-scale pyramid in transformer for video-based pedestrian retrieval. Different layers employ different-direction division strategies. After the process of transformer layer, part-informed features are extracted. Each layer contains features with different scales. The four layers have 1, $D_v$, $D_h$, and $D_p$ features, respectively.}
	\label{motivation}
\end{figure}

Recently, the transformer structure has achieved incredible progress in computer vision. The transformer structure is a popular model in natural language processing (NLP) and can handle the sequence data effectively. In computer vision, the input image is split into many patches. These patches are regarded as tokens similar to the words in the NLP task. A sequence of feature embedding of these patches is fed to the transformer layers. With the help of the multi-head self-attention module, the transformer can obtain global-level relationships among all the patches without losing information. Meanwhile, the CNN convolution kernel can only perceive limited scope, and the down-sampling operation inevitably loses much information. Moreover, the transformer structure has achieved competitive performance compared with CNN \cite{RN710}. Although the transformer has a global perception, the fine-grained information may be neglected, which results in a limited performance.

\IEEEpubidadjcol This paper proposes a multi-direction and multi-scale Pyramid in Transformer (PiT) for video pedestrian retrieval, as illustrated in Fig. \ref{motivation}. The pyramid contains four layers, which adopt ``no division'', vertical, horizontal, and patch-based division strategies, respectively. The first layer includes a global-level feature of the pedestrian image. The second, third, and fourth layers contain $D_v$, $D_h$, and $D_p$ part-level features. In this way, the proposed PiT applies multi-direction division strategies in the pedestrian image and extracts a multi-scale feature representation for each pedestrian image. 

Concretely, each pedestrian image is split into many patches. A class token and the feature embeddings of all patches are flattened and fed to multiple transformer layers. Then the processed patch tokens are rearranged into a two-dimension structure according to their original positions. Different division strategies are applied to this two-dimension structure, which generates different-direction parts. The class token and patch tokens within the same part are flattened to form a new token sequence. After the process of the last transformer layer, the class token learns the multi-direction part-informed information. A feature pyramid for each pedestrian image is obtained by combining all the features with different scales from different layers. The corresponding features of all the images within the same video are fused to generate the final multi-direction and multi-scale feature pyramid.

The traditional CNN-based methods usually apply horizontal division to feature map \cite{RN191}, which is reasonable because each horizontal stripe usually contains the head, torso, or legs. However, the vertical division can divide the human body into the right limb, head and torso, and the left limb, which introduces part-informed clues with more dimensions. Applying vertical and horizontal division simultaneously, which forms the patch-based division, can also provide more fine-grained information. Combining these multi-direction and multi-scale features can effectively improve the model performance.  
Experiments on two challenging video-based benchmarks, MARS \cite{RN158} and iLIDS-VID \cite{RN386}, show the proposed PiT achieves state-of-the-art performance. Extensive ablation studies also demonstrate the superiority of the proposed pyramid structure. 

The main contributions of this paper can be summarized as follows:
\begin{itemize}
	\item \textbf{Multi-direction}: the proposed vertical and horizontal division strategies in transformer introduce fine-grained, part-informed information from different directions. 
	\item \textbf{Multi-scale}: the global and local-level features with different scales form a feature pyramid. This multi-scale combination makes the feature representation rich and discriminative.
	\item \textbf{Performance}: the proposed PiT achieves state-of-the-art performance on two challenging video-based benchmarks, and extensive ablation studies demonstrate the superiority of the proposed multi-direction and multi-scale pyramid structure.
\end{itemize}

The rest of this paper is organized as follows: the related works are reviewed and analyzed in Section \ref{related work}, and then the proposed method is introduced in Section \ref{method}. Experimental results and analysis are presented in Section \ref{Experiments}, and Section \ref{conlusion} concludes this paper.

\section{Related Work} \label{related work}

\subsection{Video-based Pedestrian Retrieval}
An early method for image-based pedestrian retrieval fused various features, such as RGB, HSV, HoG, LOMO, etc., to achieve the multi-feature fusion and overcome the challenge from pedestrian appearance changes \cite{RN708}. Whereas video-based pedestrian retrieval has multiple images for each pedestrian. These consecutive pedestrian images can provide abundant temporal and spatial information, which can alleviate the negative effects of appearance change, occlusion, pose variation, etc \cite{RN686}. Therefore, existing methods focus on exploiting both spatial and temporal clues from pedestrian video. GRL \cite{RN606} employ video-level features to guide the generation of correlation map and disentangle the frame-level features into high-correlation and low-correlation features. BiCnet-TKS \cite{RN607} introduced a bilateral complementary network to mine the divergent body parts of each pedestrian and proposed a temporal kernel selection module to explore temporal relations adaptively. 
CTL \cite{RN608} employed a key-point estimator to extract multi-scale semantic features to form a topology graph. Then a 3D graph convolution is used to capture hierarchical spatial and temporal dependencies. 
Besides, AGW$_+$ \cite{RN709} employed a frame-level average pooling for video feature representation, which is simple but effective.

These methods above mainly utilized CNN to extract spatial and temporal clues. However, the CNN convolution kernel cannot capture long-range relationships, and the CNN down-sampling operation results in inevitable information loss.

\subsection{Vision Transformer}
Recently, transformer architecture has become a de-facto standard for natural language processing. ViT \cite{RN710} introduced this architecture to computer vision and achieved better performance than many state-of-the-art methods on the image classification task. Following this improvement, many works were proposed to improve the performance of the transformer-based framework. For the video-based classification task, Swin transformer \cite{RN630} proposed a hierarchical structure and utilized shifted windows to solve the non-overlapping patch division problem. A 3D shifted window is also proposed to preprocess video data.

These methods above demonstrated the transformer structure could perform well for the video classification task. However, the video-based pedestrian retrieval is very different from the video classification task. The video-based pedestrian retrieval task depends highly on appearance information rather than motion information. Therefore, the transformer structure with the ability to perceive more fine-grained information is needed for the video-based pedestrian retrieval task.

\begin{figure*}[]
	\centering
	\includegraphics[width=1\textwidth]{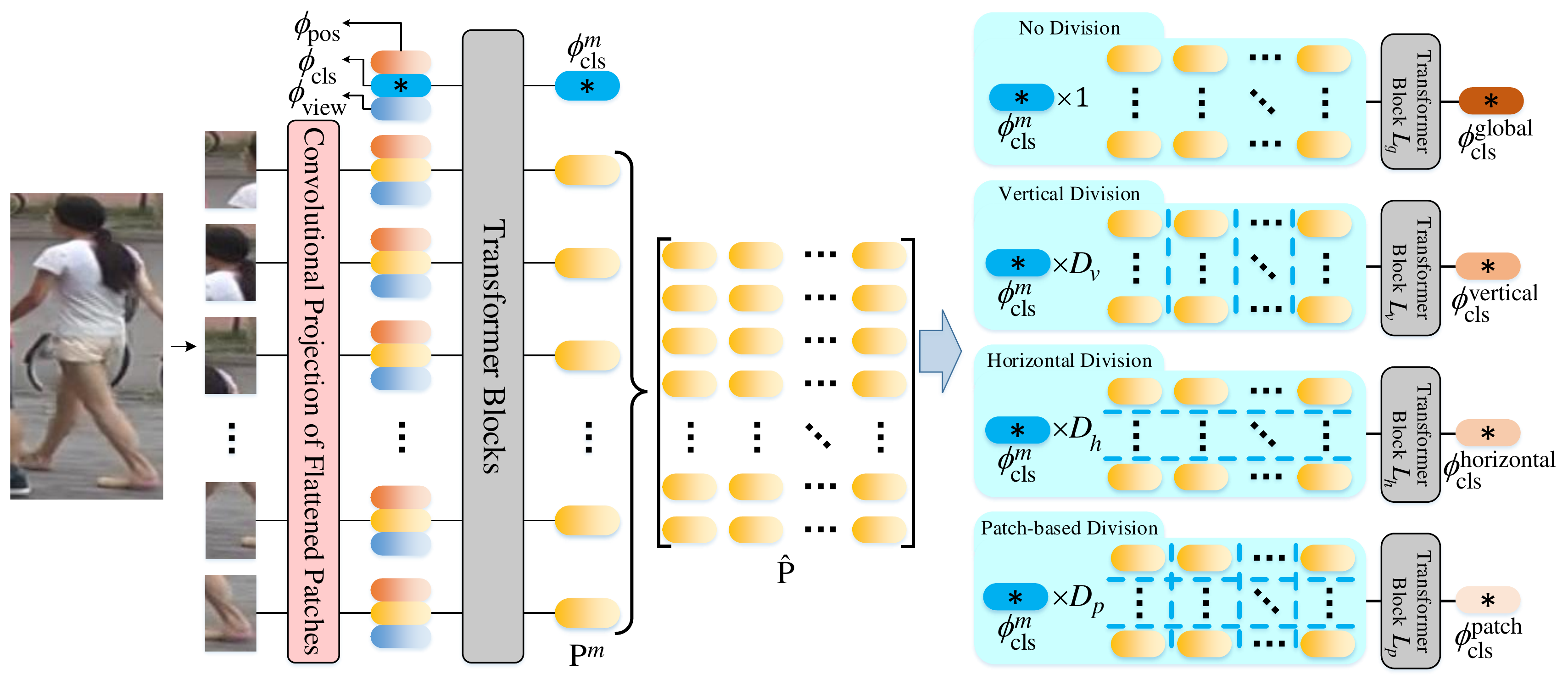}
	\caption{The proposed feature Pyramid in Transformer (PiT) for each pedestrian image. The feature pyramid with four layers contains multi-direction and multi-scale feature representations.}
	\label{framework}
\end{figure*}

\subsection{Division Strategies for Pedestrian Retrieval}
For the pedestrian retrieval task, dividing the feature map into multiple stripes is a typical operation. PCB \cite{RN191} proposed to horizontally divide the feature map into multiple stripes and achieved significant performance improvement compared with the original model. 
SPP \cite{RN570} divided the feature map into equal patches and assembled the multi-scale feature patches into a pyramid structure. 
HPP \cite{RN318} employed the horizontal division to form a multi-scale pyramid and improved the model performance for this task.

Although the models above utilized different division strategies, performance comparison among different division strategies in a unified framework has not been conducted, thus the differences between these strategies remain to be explored.

\section{Multi-direction and Multi-scale Pyramid in Transformer} \label{method}

\subsection{Transformer-based Framework} \label{transformer}
The Vision Transformer (ViT) \cite{RN710} is employed to construct the framework, as illustrated in Fig. \ref{framework}. Given some pedestrian videos $\{V_1, V_2, \cdots\}$ and pedestrian IDs $\{y_1, y_2, \cdots\}$, each video $V$ contains $K$ pedestrian images $I$, as $V = \{I_1, I_2, \cdots, I_K\}$. 
A convolution layer is used to embed the pedestrian image into multiple feature embeddings. Concretely, after applying convolution operation on the pedestrian image, a feature map $f \in \mathbb{R}^{h \times w \times c}$ is obtained. Then the feature map is flattened to generate $N$ features, where $N = h \cdot w$ and the size of each feature is $1\times c$. In this way, each feature can be treated as the feature embedding of each image patch, and the size of each image patch is the same as convolution kernel size $k$. The convolution stride $s$ determines the interval of the adjacent image patches.

The feature embedding of each image patch from the image $I$ is also called patch token $p$. A class token $\phi_\text{cls}$ with the size of $1 \times c$ is also introduced to represent the feature embedding of the whole image. After flattening class and patch tokens into a sequence $\{\phi_\text{cls}; p_1; \dots; p_N\}$, the patch tokens lose the location information in original pedestrian image. Therefore, a position embedding $\phi_\text{pos} \in \mathbb{R}^{(N+1)\times c}$ is employed to retain this location information. 
A camera embedding $\hat{\phi}_\text{view}$ is also introduced to keep the camera information. 
Since the class token and all the patch tokens belong to the same camera, the size of $\hat{\phi}_\text{view}$ is set to $1\times c$. Then $\hat{\phi}_\text{view}$ is copied $N+1$ times to form a new embedding $\phi_\text{view} \in \mathbb{R}^{(N+1)\times c}$. 
After these operations, the token sequence $z^0$ from the pedestrian image $I$ is calculated as follows,
\begin{equation}
	\begin{split}
		z^0 & = [\phi_\text{cls}; p_1; \dots; p_N] + \lambda_1 \phi_\text{pos} + \lambda_2 \phi_\text{view} \\
		& = [\phi_\text{cls}^0; p_1^0; \cdots; p_N^0], \label{X0}
	\end{split} 
\end{equation}
where $\lambda_1$ and $\lambda_2$ are trade-off parameters. Then $m$ transformer layers are employed to learn the relationship between the class token and all patch tokens. Each transformer layer is composed of a Multi-head Self-Attention (\texttt{MSA}) block and a Multi-Layer Perception (\texttt{MLP}) block. A LayerNorm (\texttt{LN}) layer is applied before \texttt{MSA} and \texttt{MLP} blocks, and a shortcut connection is also employed as follows,
\begin{equation}
	\begin{split}
		z'  &= z^{d-1} + \texttt{MSA}(\texttt{LN}(z^{d-1})),    \\
		z^d &= z' + \texttt{MLP}(\texttt{LN}(z')). \label{MSA}
	\end{split}
\end{equation}

After $m$ transformer layers, the vector sequence $z^m$ is obtained.
\begin{equation}
	z^m = [\phi_\text{cls}^m; p_1^m; p_2^m; \cdots; p_N^m] := [\phi_\text{cls}^m; \mathrm{P}^m]. \label{XM}
\end{equation}
In Eq. \ref{XM}, $N$ patch tokens $\{p_i^m\}_{i=1}^N$ 
are denoted as $\mathrm{P}^m$,
\begin{equation}
	\mathrm{P}^m = [p_1^m; p_2^m; \cdots; p_N^m]. \label{pm}
\end{equation}

\subsection{Multi-direction and Multi-scale Pyramid} \label{pyramid}
To explore the fine-grained, part-informed information, the patch tokens $\mathrm{P}^m$ are rearranged into a new form $\widehat{\mathrm{P}}$
according to their original positions in pedestrian image $I$,
\begin{equation}
	\setlength{\belowdisplayskip}{18pt}
	\underbrace{[p_1^m; p_2^m; \cdots; p_N^m]}_{\mathrm{P}^m\in \mathbb{R}^{N\times c}} \xrightarrow[]{\text{rearrange}} \underbrace{\left[
		\begin{matrix}
			p_1^m & p_2^m & \cdots & p_w^m \\
			p_{w+1}^m & p_{w+2}^m & \cdots & p_{2w}^m \\
			\vdots      & \vdots      & \ddots & \vdots      \\
			\cdots      & \cdots      & \cdots & p_{N}^m      \\
		\end{matrix}
		\right]}_{\widehat{\mathrm{P}}\in \mathbb{R}^{h\times w \times c}}. \label{Xpatch}
\end{equation}

The rearranged patch tokens $\widehat{\mathrm{P}}$ have the same size as the feature map $f\in \mathbb{R}^{h\times w \times c}$. We learn from the division strategies in Convolution Neural Network (CNN) and apply similar strategies to patch tokens $\widehat{\mathrm{P}}$.

\subsubsection{\textbf{Multi-direction} Division Strategies} 
The rearranged tokens $\widehat{\mathrm{P}}$ are copied four times. Multi-direction division strategies are applied to these four copies.

For the first copy, ``no division'' is applied to the tokes $\widehat{\mathrm{P}}$. Then the class token $\phi_\text{cls}^m$ along with $\widehat{\mathrm{P}}$ are flattened to a new token sequence $z_\text{global} \in \mathbb{R}^{(N+1)\times c}$, which equals $z^m$.
\begin{equation}
	z_\text{global} = [\phi_\text{cls}^m; p_1^m; p_2^m; \cdots; p_N^m] \Leftrightarrow z^m. \label{zglobal}
\end{equation}
The transformer layer $L_{g}$ receives this sequence and outputs a new sequence $z'_\text{global} \in \mathbb{R}^{(N+1)\times c}$. 
We follow the general operation in \cite{RN710} \cite{RN630}, which discard all the patch tokens and only keep the class token for the following operations. Therefore, only the class token $\phi_\text{cls}^\text{global} \in \mathbb{R}^{1\times c}$ is kept as the feature representation of the whole pedestrian image $I$.

For the second copy, ``vertical division'' is applied to the patch tokens $\widehat{\mathrm{P}}$ along the vertical direction to generate $D_v$ parts. 
Each part has $N/D_v$ patch tokens. The class token is copied $D_v$ times, and each one is assigned to one part. Then the class token and all the patch tokens in the corresponding part are flattened along the vertical direction to form a new vector sequence $z_\text{vertical} \in \mathbb{R}^{(N/D_v+1)\times c}$. 
There are $D_v$ sequences in total.
For example, the first sequence $z_{\text{vertical},1}$ is shown as follows,
\begin{equation}
	z_{\text{vertical},1} = [\phi_\text{cls}^m; p_1^m; p_{w+1}^m; p_{2w+1}^m;\cdots]. \label{zvertical}
\end{equation}
Each sequence $z_{\text{vertical},i}$ is followed by the parameter-sharing transformer layer $L_{v}$.
In this way, the relationship between the class token and a specific vertical part is explored. 
After the process of the transformer layer $L_{v}$, $D_v$ class tokens $\{\phi_{\text{cls},i}^\text{vertical}\}_{i=1}^{D_v}$ are kept and denoted as $\phi_\text{cls}^\text{vertical}$. Moreover, this paper first proposes the ``vertical division'' strategy, which extracts fine-grained feature representation and also introduces significant performance improvement.

\begin{figure}[htp]
	\centering
	\includegraphics[width=0.48\textwidth]{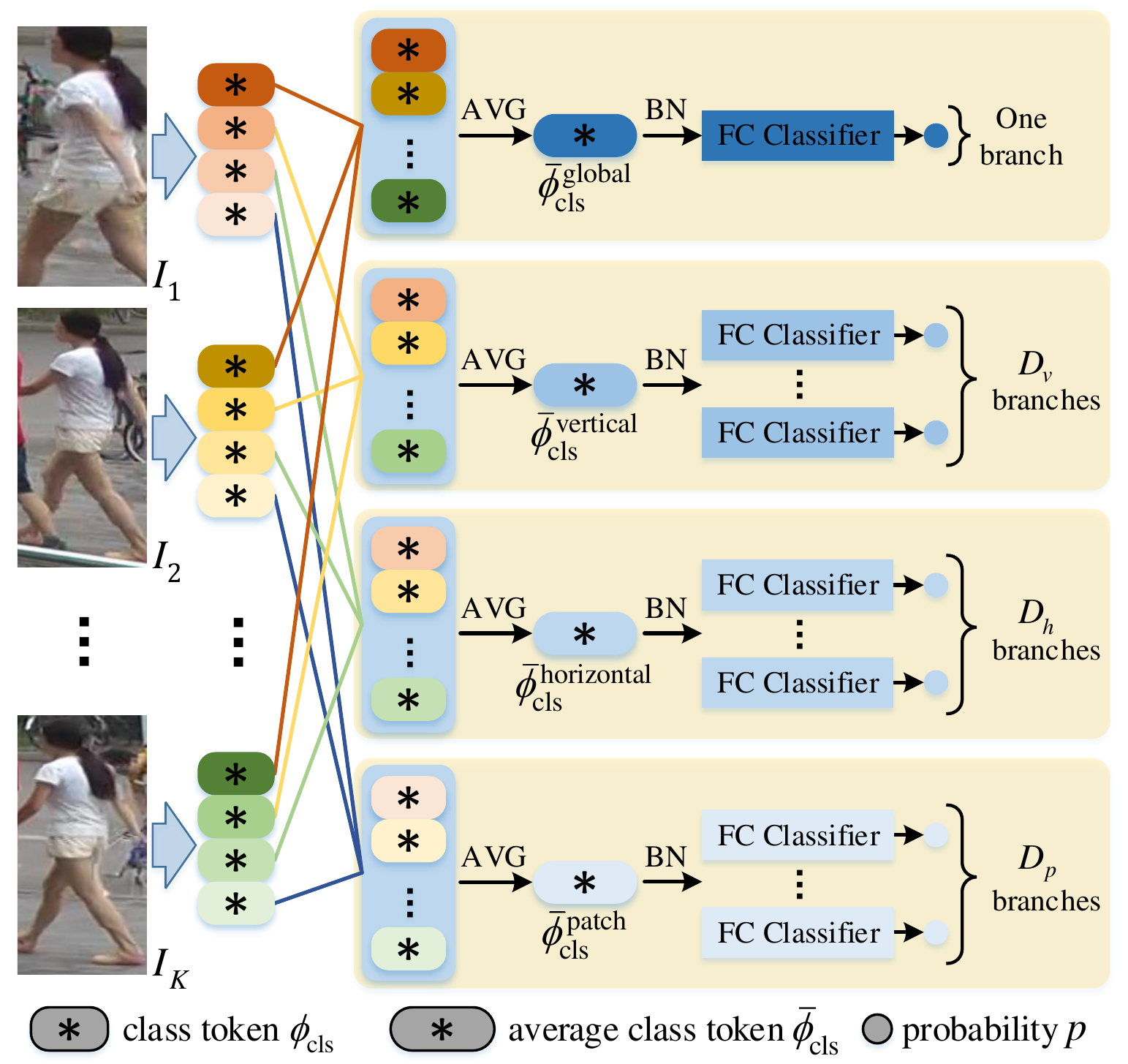}
	\caption{The generation and training process of the feature pyramid for each video. The corresponding features of each pedestrian image are averaged to generate the final feature, and each feature is trained individually.} \label{horizontal}
\end{figure}

For the third copy, ``horizontal division'' is applied by dividing the patch tokens $\widehat{\mathrm{P}}$ into $D_h$ parts. 
Horizontal division strategy is applied along the horizontal direction, and each part has $N/D_h$ patch tokens. After the division operation, the class token is copied $D_h$ times. Each one and its corresponding patch tokens are flattened along the horizontal direction to form the token sequence $z_\text{horizontal} \in \mathbb{R}^{(N/D_h+1)\times c}$.
The first sequence $z_{\text{horizontal},1}$ is shown below as an example,
\begin{equation}
	z_{\text{horizontal},1} = [\phi_\text{cls}^m; p_1^m; p_2^m; p_3^m; \cdots; p_{N/D_h}^m]. \label{zvertical}
\end{equation}
All the sequences $\{z_{\text{horizontal},i}\}_{i=1}^{D_h}$ are followed by the parameter-sharing transformer layers $L_{h}$, and $D_h$ class tokens $\{\phi_{\text{cls},i}^\text{horizontal}\}_{i=1}^{D_h}$ are obtained and denoted as $\phi_\text{cls}^\text{horizontal}$. The ``horizontal division'' is proposed for CNN in previous work \cite{RN318}. However, we propose a manner to apply this strategy to the new structure, \textit{i.e.,} vision transformer, which is proven to be effective. 

For the fourth copy, the vertical and horizontal division strategies are applied simultaneously to the rearranged patch tokens $\widehat{\mathrm{P}}$ to form ``patch-based division''. Each part has $N/D_p$ patch tokens and has a more fine-grained receptive field.
After the division operation, the class token and all patch tokens in the corresponding part are flattened along the horizontal direction to form $D_p$ token sequences $z_\text{patch} \in \mathbb{R}^{(N/D_p+1)\times c}$, where $D_p = D_v\times D_h$. 
The first sequence $z_{\text{patch},1}$ is shown below as an example,
\begin{equation}
	z_{\text{patch},1} = [\phi_\text{cls}^m; p_1^m; p_2^m; \cdots; p_{N/D_v}^m;p_{w+1}^m; p_{w+2}^m;\cdots]. \label{zvertical}
\end{equation}

Each token sequence $z_{\text{patch},i}$ is 
followed by the transformer layers $L_{p}$, which generate $D_p$ class tokens $\{\phi_{\text{cls},i}^\text{patch}\}_{i=1}^{D_p}$. These class tokens are denoted as $\phi_\text{cls}^\text{patch}$. A similar patch-based division strategy was proposed in CNN structure \cite{RN570}. However, this paper proposes multi-direction division strategies, which are very different from the previous model.

After the transformer layer, each class token 
has the perception within its corresponding global/vertical/horizontal/patch-based part, making it obtain more fine-grained local information. 

\subsubsection{\textbf{Multi-scale} Pyramid Structure}
The rearranged $\widehat{\mathrm{P}}$ is divided using different scales, which generates multi-scale feature representations. These features are concatenated to form a pyramid structure $z_\phi\in\mathbb{R}^{(1+D_v+D_h+D_p)\times c}$ for each pedestrian image $I$ as follows,
\begin{equation}
	z_\phi = [\phi_\text{cls}^\text{global};\phi_\text{cls}^\text{vertical};
	\phi_\text{cls}^\text{horizontal};\phi_\text{cls}^\text{patch}], \label{Xphi}
\end{equation}
where $\phi_\text{cls}^\text{global}$, $\phi_\text{cls}^\text{vertical}$, $\phi_\text{cls}^\text{horizontal}$, and $\phi_\text{cls}^\text{patch}$ are in the space of $\mathbb{R}^{1\times c}$, $\mathbb{R}^{D_v\times c}$, $\mathbb{R}^{D_h\times c}$, and $\mathbb{R}^{D_p\times c}$, respectively. 

There are $K$ images in each pedestrian video $V$, and each image is represented by $z_\phi$.
The feature pyramid $\bar{z}_\phi$ of the video $V$ is generated as illustrated in Fig. \ref{horizontal}. Each feature of the video $V$ is the corresponding average feature within it. For example, the class token $\bar{\phi}_\text{cls}^\text{global}$ is calculated as follows,
\begin{equation}
	\bar{\phi}_\text{cls}^\text{global} = \sum_{k=1}^K \phi_{\text{cls}, k}^\text{global}. \label{barphi}
\end{equation}

The feature pyramid $\bar{z}_\phi\in\mathbb{R}^{(1\mathcal{+}D_v\mathcal{+}D_h\mathcal{+}D_p)\times c}$ of video $V$ is expressed as follow,
\begin{equation}
	\bar{z}_\phi = [\bar{\phi}_\text{cls}^\text{global};\bar{\phi}_\text{cls}^\text{vertical};
	\bar{\phi}_\text{cls}^\text{horizontal};\bar{\phi}_\text{cls}^\text{patch}]. \label{XXphi}
\end{equation}

The CNN-based methods usually design various fusion strategies to combine pedestrian image features within the same video. The feature map from CNN contains rich spatial and temporal information, and a sophisticated fusion strategy can effectively combine these features.
However, in a transformer-based framework, the class token is not generated from the input image directly. In other words, these class tokens do not contain spatial and temporal information explicitly. Therefore, each feature of the video $V$ is obtained in an average manner as Eq. \ref{barphi}.

The multi-scale pyramid structure $\bar{z}_\phi\in\mathbb{R}^{(1\mathcal{+}D_v\mathcal{+}D_h\mathcal{+}D_p)\times c}$ contains part-informed information of different scales, which is more rich and discriminative than the original global-level feature representation. The ablation study in the latter section demonstrates the superiority of this multi-scale pyramid structure. 

\subsection{Model Training and Testing} \label{train and test}
This section describes the details of model training and testing.
To simplify the expression, the class token $\phi^\theta_\text{cls} \in \mathbb{R}^{1\times c}$ is used to represent each element in $\bar{z}_\phi$. Each class token $\phi^\theta_\text{cls}$ is followed by a \texttt{BatchNorm} layer and a classifier layer to generate the final probability $p^\theta$, as illustrated in Fig. \ref{horizontal}. There are $1\mathcal{+}D_v\mathcal{+}D_h\mathcal{+}D_p$ classifier layers to train each token $\phi^\theta_\text{cls}$ individually. The classification loss $\mathcal{L}_{cls}$ and triplet loss $\mathcal{L}_{tri}$ are used to supervise this model by averaging the losses of these independent branches.
\begin{equation}
	\mathcal{L}_{cls} = -\frac{1}{TN_c}\sum_{i=1}^{T}\sum_{j=1}^{N_c} y_j\log p_{i,j}^{\theta}, \label{Lcls}
\end{equation}
\begin{equation}
	\begin{split}
		\mathcal{L}_{tri} = \frac{1}{BT}\sum_{i=1}^{T}\sum_{a\in b_i} \ln \{1 \mathcal{+}\exp [\max d(\phi_{i,a}^{\theta},\phi_{i,p}^{\theta}) \\
		\mathcal{-}\min d(\phi_{i,a}^{\theta},\phi_{i,n}^{\theta})] \} , \label{Ltri}
	\end{split}
\end{equation}
where $T\mathcal{=}1\mathcal{+}D_v\mathcal{+}D_h\mathcal{+}D_p$, $N_c$ is the class number for a specific benchmark, $b_i$ represents the $\text{i}^{\text{th}}$ mini-batch, $B$ is the number of pedestrian images in this mini-batch, $a,p,n$ are anchor, positive, negative samples, respectively. Function $d(\cdot)$ calculates the Euclidean distance between two features. The overall loss function $\mathcal{L}$ is calculated as follows,
\begin{equation}
	\mathcal{L} = \mathcal{L}_{cls} + \mathcal{L}_{tri}. \label{L}
\end{equation} 

\begin{table*}[htp]
	\centering
	\begin{threeparttable}
		\centering
		\caption{Performance comparisons between proposed PiT and state-of-the-art methods on MARS and iLIDS-VID.} \label{SOTA}
		\begin{tabular}{lclllllllll}
			\hline
			\multicolumn{1}{c}{\multirow{2.8}{*}{Methods}} & \multirow{2.8}{*}{Venue} & \multicolumn{4}{c}{MARS}              & \multicolumn{4}{c}{iLIDS-VID}  \\
			\cmidrule(r){3-6} \cmidrule(r){7-10}
			\multicolumn{1}{c}{}                         &                        & Rank-1    & Rank-5    & Rank-10    & mAP   & Rank-1    & Rank-5    & Rank-10   & Rank-20    \\ \hline
			ADFD        \cite{RN421}                               & CVPR2019               & 87.00 & 95.40 &        & 78.20 & 86.30 & 97.40 &       & 99.70  \\
			GLTR        \cite{RN422}                               & ICCV2019               & 87.02 & 95.76 &        & 78.47 & 86.00 & 98.00 &       &        \\
			COSAM       \cite{RN423}                               & ICCV2019               & 84.90 & 95.50 &        & 79.90 & 79.60 & 95.30 &       &        \\
			AGW$_+$ \cite{RN709}  & TPAMI2020   & 87.60 &  &  & 83.00 & 83.20 & 98.30 &       &        \\
			RTF         \cite{RN558}                               & AAAI2020               & 87.10 &       &        & 85.20 & 87.70 &       &       &        \\
			FGRA        \cite{RN616}                               & AAAI2020               & 87.30 & 96.00 &        & 81.20 & 88.00 & 96.70 & 98.00 & 99.30  \\
			RGSATR      \cite{RN617}                               & AAAI2020               & 89.40 & 96.90 &        & 84.00 & 86.00 & 98.00 &       & 99.40  \\
			AMEM        \cite{RN619}                               & AAAI2020               & 86.70 & 94.00 &        & 79.30 & 87.20 & 97.70 &       & 99.50  \\
			CSTNet      \cite{RN567}                               & IJCAI2020              & 90.20 & 96.80 &        & 83.90 & 87.80 & 98.50 &       & 99.60  \\
			ASTA-Net    \cite{RN615}                               & ACM MM2020             & 90.40 & 97.00 &        & 84.10 & 88.10 & 98.60 &       &        \\
			MG-RAFA     \cite{RN418}                               & CVPR2020               & 88.80 & 97.00 &        & 85.90 & 88.60 & 98.00 &       & 99.70  \\
			MGH         \cite{RN424}                               & CVPR2020               & 90.00 & 96.70 &        & 85.80 & 85.60 & 97.10 &       & 99.50  \\
			STGCN       \cite{RN425}                               & CVPR2020               & 89.95 & 96.41 &        & 83.70 &       &       &       &        \\
			VRSTC       \cite{RN441}                               & CVPR2020               & 88.50 & 96.50 & 97.40  & 82.30 & 83.40 & 95.50 & 97.70 & 99.50  \\
			TCLNet-tri* \cite{RN614}                               & ECCV2020               & 89.80 &       &        & 85.10 & 86.60 &       &       &        \\
			AP3D        \cite{RN618}                               & ECCV2020               & 90.70 &       &        & 85.60 & 88.70 &       &       &        \\
			AFA         \cite{RN565}                               & ECCV2020               & 90.20 & 96.60 &        & 82.90 & 88.50 & 96.80 &       & 99.70  \\
			SSN3D       \cite{RN620}                               & AAAI2021               & 90.10 & 96.60 & 98.00  & 86.20 & 88.90 & 97.30 &       & 98.80  \\
			GRL         \cite{RN606}                               & CVPR2021               & 91.00 & 96.70 &        & 84.80 & 90.40 & 98.30 &       & 99.80  \\
			BiCnet-TKS  \cite{RN607}                               & CVPR2021               & 90.20 &       &        & 86.00 &       &       &       &        \\
			CTL         \cite{RN608}                               & CVPR2021               & \textbf{91.40} & 96.80 &  & 86.70 & 89.70 & 97.00 &       & 100.00 \\ \hline
			\multicolumn{2}{c}{Proposed PiT}                                              & 90.22 & \textbf{97.23} & \textbf{98.04} & \textbf{86.80} & \textbf{92.07} & \textbf{98.93} & \textbf{99.80} & \textbf{100.00} \\ \hline
		\end{tabular}
		\begin{tablenotes}
			\item[1] The best results are in bold. 
		\end{tablenotes}
	\end{threeparttable}
\end{table*}

In the testing process, the multi-direction and multi-scale feature pyramid $\bar{z}_\phi \in\mathbb{R}^{(1\mathcal{+}D_v\mathcal{+}D_h\mathcal{+}D_p)\times c}$ is used to represent the feature representation of pedestrian video $V$.

\section{Experiments} \label{Experiments}
\subsection{Experimental Setting}
\subsubsection{Benchmarks}
The proposed Pyramid in Transformer (PiT) is evaluated in two challenging benchmarks: MARS \cite{RN158} and iLIDS-VID \cite{RN386}. There are also two other popular benchmarks: DukeMTMC-VideoReID and PRID2011. Existing methods \cite{RN606} \cite{RN607} \cite{RN620} have achieved more than 0.95 in terms of Rank-1 metric on these two benchmarks. However, there is still much room for improvement on the challenging MARS and iLIDS-VID benchmarks.  
\begin{itemize}
	\item MARS is the largest video-based pedestrian retrieval benchmark and captured by six cameras on a university campus. It contains 20,478 videos from 1,261 identities. These videos are generated by employing the DPM detector and GMMCP tracker, which results in many videos with poor qualities.
	\item iLIDS-VID is captured by two cameras in an airport hall. It contains 600 videos from 300 identities. This benchmark is very challenging due to pervasive background clutter, mutual occlusions, and lighting variations. 
\end{itemize}

\subsubsection{Evaluation Protocol and Metrics}
For MARS benchmark, the standard training and testing split provided by \cite{RN158} is used for training the proposed PiT. The Cumulative Matching Characteristic (CMC) curve and mean Average Precision (mAP) are employed for evaluation. For iLIDS-VID benchmark, the whole set of videos is randomly divided into two halves. Then the trials are repeated ten times, and the CMC curve is used to evaluate the average results. 
For convenience, Rank-1, Rank-5, Rank-10, and Rank-20 are employed to represent the CMC curve.

\subsubsection{Implementation Details} 
The proposed PiT is implemented using Pytorch. The transformer ViT-B16 \cite{RN710}, pre-trained on ImageNet, is employed as the backbone. We follow the operation in \cite{RN711} to preprocess all the videos. Specifically, each video is divided equally into $K$ snippets, where $K$ equals 8. The first pedestrian image in each snippet is selected as the keyframe, and $K$ images are used to represent this video. Each pedestrian image is resized to $256\times 128$. The batch size and parameter $m$ are set to 16 and 11. The kernel size $k$ and stride $s$ of the convolution layer are set to 16 and 12. The dimension $h\times w \times c$ of feature embedding outputted by convolution layer is $21\times 10\times 768$. The trade-off parameters $\lambda_1$ and $\lambda_2$ are set to $1.0$ and $1.5$. The division parameters $D_v$, $D_h$, $D_p$ are 2, 3, and 6. The standard Stochastic Gradient Descent (SGD) with momentum and an initial learning rate of 0.01 is used to train these models 120 epochs for each benchmark. Cosine annealing is employed to schedule the learning rate. The convolution layer and transformer layers are frozen in the first five epochs to train the classifier layers. After these five epochs, the whole network is trained.

\subsection{Comparison with State-of-the-art Methods}
Table \ref{SOTA} shows the comparison between the proposed PiT and twenty other state-of-the-art methods 
in terms of mAP score and CMC accuracy. These state-of-the-art methods are all within three years and employed ResNet50 as their backbone to explore the spatial and temporal information among pedestrian images. They used attribute information \cite{RN421} \cite{RN619}, attention mechanism \cite{RN423} \cite{RN418}, graph convolution \cite{RN424} \cite{RN425} \cite{RN608}, 3D convolution \cite{RN618} \cite{RN620}, relation-guided models \cite{RN616} \cite{RN617} \cite{RN606}, Generative Adversarial Networks (GAN) \cite{RN441}, and new network architectures \cite{RN709} \cite{RN558} \cite{RN567} \cite{RN615} \cite{RN614} \cite{RN565} \cite{RN422} \cite{RN607}, respectively, to generate the feature representation of each pedestrian video. Meanwhile, the proposed PiT employs a transformer-based framework and utilizes the simple average fusion to obtain the multi-direction and multi-scale feature pyramid.

\subsubsection{Performances on MARS} 
Compared with other state-of-the-art methods, the proposed PiT achieves the best mAP score and competitive CMC accuracy. The best competitor, CTL \cite{RN608}, utilized a key-points estimator to extract human body local features as graph nodes and achieved topology learning for video-based pedestrian retrieval. In comparison, the proposed PiT does not explore the relationship among different pedestrian images within the same video and reaches a better mAP value. This demonstrates the proposed feature pyramid containing more fine-grained local information has a better generalization performance.

\begin{table}[htp]
	\centering
	\caption{Performance comparisons between different-direction division strategies. 
	} \label{strategies}
	\begin{tabular}{clccc} \hline
		\multicolumn{1}{c}{\multirow{2.5}{*}{One layer}}                    & \multicolumn{1}{c}{\multirow{2.5}{*}{Parameter}}   & \multicolumn{2}{c}{MARS}  &   iLIDS-VID     \\
		\cmidrule(r){3-4} \cmidrule(r){5-5}
		\multicolumn{1}{l}{}                    &  & Rank-1  & mAP  & Rank-1 \\ \hline
		\multicolumn{1}{c}{No Division (Baseline)}            & $1\mathcal{\times}210$ & 87.33   & 84.00 & 89.87  \\ \hline
		\multirow{5}{*}{Vertical Division}      
		& $105\mathtt{\times}2$ & 88.26  & 85.07 & 89.73\\
		& $70\mathtt{\times}3$  & 88.64  & 85.72 & 90.60\\
		& $42\mathtt{\times}5$  & 88.80  & 85.56 & \textbf{90.93}\\
		& $35\mathtt{\times}6$  & 89.08  & 85.96 & 90.40\\
		& $30\mathtt{\times}7$  & \textbf{89.78}  & \textbf{85.99} & 89.93\\ \hline
		\multirow{5}{*}{Horizontal Division}    
		& $2\mathtt{\times}105$ & 88.26   & 85.33 & 90.07\\
		& $3\mathtt{\times}70$  & 89.35   & 85.86 & 90.20\\
		& $5\mathtt{\times}42$  & \textbf{89.46}   & \textbf{86.24} & \textbf{91.40} \\
		& $6\mathtt{\times}35$  & 89.18   & 85.94 & 90.73 \\
		& $7\mathtt{\times}30$  & 89.35   & 86.00 & 90.67 \\ \hline
		\multirow{3}{*}{Patch-based Division} 
		& $6\mathtt{p}$ & 89.13   & 86.01 & 91.00\\
		& $14\mathtt{p}$   & 89.24  & 86.11 & \textbf{91.67} \\
		& $15\mathtt{p}$   & \textbf{89.73}   & \textbf{86.17} & 90.80\\ \hline
	\end{tabular}
\end{table}

\subsubsection{Performances on iLIDS-VID}
Compared with other methods, the proposed PiT achieves state-of-the-art performance. The best competitor, GRL \cite{RN606}, used global correlation estimation to disentangle features into high-correlation and low-correlation features. Then GRL proposed temporal reciprocating learning to enhance the high-correlation semantic clues and accumulate the low-correlation sub-critical clues for the final feature representation. Meanwhile, the proposed PiT has a concise network structure and achieves 1.67\% performance improvement on the metric of Rank-1 than GRL.

\begin{table*}[htp]
	\centering
	\caption{Performance comparisons between different combinations. 
	} \label{combinations}
	\begin{tabular}{cllccc} \hline
		& \multicolumn{1}{l}{\multirow{2.6}{*}{Type}} & \multicolumn{1}{l}{\multirow{2.6}{*}{Parameter}} & \multicolumn{2}{c}{MARS}        & \multicolumn{1}{c}{iLIDS-VID} \\
		\cmidrule(r){4-5} \cmidrule(r){6-6}
		& \multicolumn{1}{c}{}                      & \multicolumn{1}{c}{}                           & Rank-1         & mAP            & Rank-1                        \\ \hline
		\multirow{4}{*}{Four Layers}
		& Vertical Division                & $1\mathcal{\times}210\_105\mathcal{\times}2\_42\mathcal{\times}5\_30\mathcal{\times}7$            & 89.34          & 86.30          & 91.13                         \\
		& Horizontal Division              & $1\mathcal{\times}210\_2\mathcal{\times}105\_5\mathcal{\times}42\_7\mathcal{\times}30$            & 89.56          & 86.32          & 91.40                         \\
		& Patch-based Division                   & $1\mathcal{\times}210\_6\mathtt{p}\_14\mathtt{p}\_15\mathtt{p}$                  & 89.51          & 85.96          & 90.47                         \\
		& Proposed PiT                     & $1\mathcal{\times}210\_105\mathcal{\times}2\_3\mathcal{\times}70\_6\mathtt{p}$   & \textbf{90.22} & \textbf{86.80} & \textbf{92.07}         \\ \hline      
	\end{tabular}
\end{table*}

\subsection{Ablation Study}
Different division strategies in a unified framework are compared in this section. In this paper, each image is split into 210 patches. They are then further divided into $2, 3, 5, 6, 7$ parts. In other words, the optional values of parameters $D_v$ and $D_h$ are $2, 3, 5, 6, 7$. To explicitly show the division details, $105\mathtt{\times}2$, $70\mathtt{\times}3$, $42\mathtt{\times}5$, $35\mathtt{\times}6$, and $30\mathtt{\times}7$ are used to represent the division parameters using vertical division. $2\mathtt{\times}105$, $3\mathtt{\times}70$, $5\mathtt{\times}42$, $6\mathtt{\times}35$, and $7\mathtt{\times}30$ are denoted as division parameters using horizontal division. 
Applying $105\mathtt{\times}2$ vertical division and $3\mathtt{\times}70$ horizontal division simultaneously forms a patch-based division and generates $6$ parts. $105\mathtt{\times}2$ and $7\mathtt{\times}30$ generate $14$ parts, and $42\mathtt{\times}5$ and $3\mathtt{\times}70$ generate $15$ parts (\textit{i.e.}, the optional values of parameter $D_p$ are $6,14,15$). These patch-based division strategies are denoted as $6\mathtt{p}, 14\mathtt{p}, 15\mathtt{p}$. ``No division'' is denoted as $1\mathtt{\times}210$.

\subsubsection{Effectiveness of Different-direction Division Strategies} \label{one-layer}

This section compares different-direction division strategies in the transformer-based framework. The proposed pyramid with only one layer is used to compare the performances, as illustrated in Table \ref{strategies}. Compared with the baseline method, this table shows that using different-direction division strategies improves the performance. Two conclusions can be made. First, although horizontal division is a commonly used strategy, the vertical and patch-based division strategies can also improve performance effectively. Second, the best division strategy and the number of parts are different for different benchmarks. These conclusions demonstrate the need to adopt a strategy based on the practical scene.

\begin{figure}[htp]
	\centering
	\includegraphics[width=0.4\textwidth]{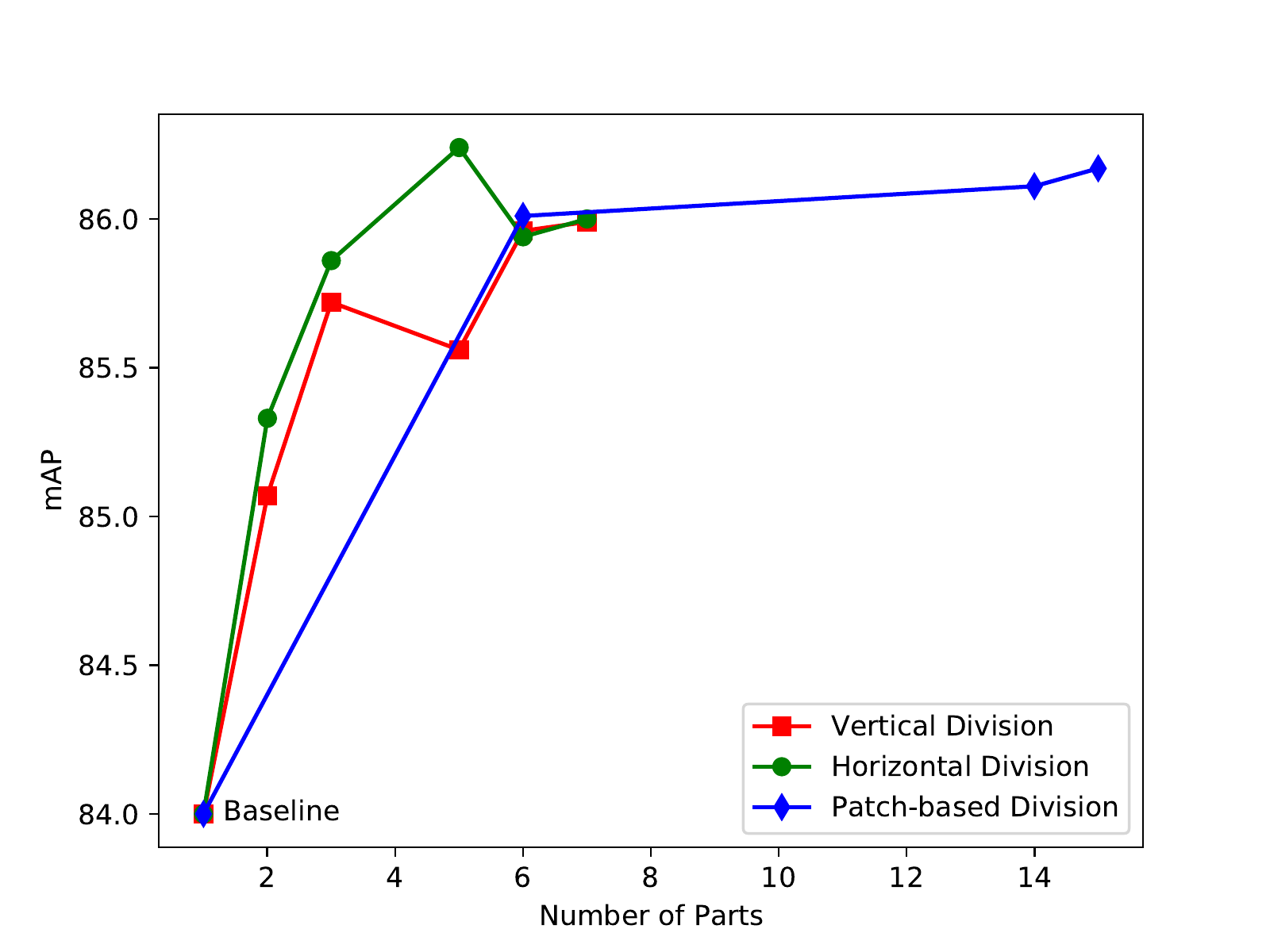}
	\caption{The mAP score changes of different-direction division strategies on MARS benchmark. The values in this figure follow the data in Table \ref{strategies}.}
	\label{MARS-1pymd}
\end{figure}

To allow for the performance analysis visually, the mAP scores of different division strategies on MARS benchmark are illustrated in Fig. \ref{MARS-1pymd}. In this figure, increasing the number of parts can improve the performance gradually. However, more number of parts also introduces more computation complexity. For different division strategies, dividing the patch tokens into six parts achieves a better trade-off between the performance and computation complexity. Therefore, the parameter of proposed PiT is $1\mathcal{\times}210\_105\mathcal{\times}2\_3\mathcal{\times}70\_6\mathtt{p}$, and its fourth layer splits the patch tokens into six parts. 

\subsubsection{Effectiveness of Multi-scale Pyramid Structure} \label{divisions}
This section shows the performances of pyramid structures with different layers, as illustrated in Table \ref{addition}. In this table, the proposed pyramid with one layer only employs ``no division''. The proposed pyramid with two layers additionally uses the vertical division strategy. Then the horizontal division and patch-based division strategies are added one by one. Each layer contains part-informed information with different scales. As the number of layers increases, the performance improves gradually. These improvements demonstrate that fusing multi-scale feature representations can improve performance effectively.

\begin{table}[tp]
	\centering
	\caption{Performance comparisons between the proposed pyramid with different numbers of layers. 
	} \label{addition}
	\begin{tabular}{clccc} \hline
		\multicolumn{1}{c}{\multirow{2.8}{*}{\begin{tabular}[c]{@{}c@{}}Number of\\Layers\end{tabular}}} & \multicolumn{1}{l}{\multirow{2.5}{*}{Parameter}} & \multicolumn{2}{c}{MARS}        & iLIDS-VID      \\
		\cmidrule(r){3-4} \cmidrule(r){5-5}
		\multicolumn{1}{c}{}                                  & \multicolumn{1}{c}{}                           & Rank-1         & mAP            & Rank-1         \\ \hline
		1                                                     & $1\mathcal{\times}210$                                                          & 87.33          & 84.00          & 89.87          \\
		2                                                     & $1\mathcal{\times}210\_105\mathcal{\times}2$                                    & 88.59          & 85.65          & 90.20          \\
		3                                                     & $1\mathcal{\times}210\_105\mathcal{\times}2\_3\mathcal{\times}70$               & 88.75          & 85.72          & 91.13          \\
		4                                                     & $1\mathcal{\times}210\_105\mathcal{\times}2\_3\mathcal{\times}70\_6\mathtt{p}$  & \textbf{90.22} & \textbf{86.80} & \textbf{92.07} \\ \hline
	\end{tabular}
\end{table} 

\subsubsection{Effectiveness of Multi-direction Pyramid Structure} \label{four-layer}
This section compares performances between multi-direction and single-direction pyramid structures, as illustrated in Table \ref{combinations}. All the pyramids in this table have four layers, and the differences between them are dependent upon which division strategy is employed. The type ``Vertical Division'' only employs vertical division strategies. The types ``Horizontal Division'' and ``Patch-based Division'' have the same meanings. 
For vertical and horizontal division strategies, the part numbers $2, 5, 7$ are chosen to form the bottom three layers. For patch-based division, $6\mathtt{p}, 14\mathtt{p}, 15\mathtt{p}$ are employed to form the bottom three layers.

Compared with other single-direction pyramids, the proposed PiT achieves the best performance. These comparisons demonstrate fusing multi-direction division strategies provides more improvement. On the other side, the type ``Horizontal Division'' performs better than the ``Vertical Division''. This shows the horizontal division strategy is more suitable for the pedestrian retrieval task. In conclusion, the proposed PiT fusing multi-direction and multi-scale feature representations is the best combination.

\begin{table}[htp]
	\centering
	\caption{Performance comparisons between proposed PiT with different parameters. 
	} \label{parameters}
	\begin{tabular}{clccc} \hline
		& \multicolumn{1}{l}{\multirow{2.6}{*}{Parameter}} & \multicolumn{2}{c}{MARS}        & \multicolumn{1}{c}{iLIDS-VID}\\ 
		\cmidrule(r){3-4} \cmidrule(r){5-5}
		& \multicolumn{1}{c}{}                           & Rank-1         & mAP            & Rank-1                        \\ \hline
		\multirow{3}{*}{\begin{tabular}[c]{@{}c@{}}Four\\Layers\end{tabular}}
		& $1\mathcal{\times}210\_105\mathcal{\times}2\_3\mathcal{\times}70\_6\mathtt{p}$                               & \textbf{90.22} & \textbf{86.80} & \textbf{92.07}                \\
		& $1\mathcal{\times}210\_105\mathcal{\times}2\_7\mathcal{\times}30\_14\mathtt{p}$                        & 89.24          & 86.13          & 91.20                         \\
		& $1\mathcal{\times}210\_42\mathcal{\times}5\_3\mathcal{\times}70\_15\mathtt{p}$                         & 89.24          & 85.82          & 90.40                         \\ \hline                
	\end{tabular}
\end{table}

\begin{table*}[htp]
	\centering
	\caption{Computation complexity and running time of the proposed PiT with different parameters.
	} \label{computation}
	\begin{tabular}{clcccccc} \hline
		\multicolumn{1}{c}{\multirow{3}{*}{\begin{tabular}[c]{@{}c@{}}Number of\\Layers\end{tabular}}} 
		& \multicolumn{1}{l}{\multirow{3}{*}{Parameter}} 
		& \multicolumn{1}{c}{\multirow{3}{*}{MACs}} 
		& \multicolumn{1}{c}{\multirow{3}{*}{\begin{tabular}[c]{@{}c@{}}Trainable\\Parameters\end{tabular}}} 
		& \multicolumn{4}{c}{Using One 24G NVIDIA TITAN RTX GPU}                          \\
		\cmidrule(r){5-8}
		\multicolumn{1}{c}{} & \multicolumn{1}{c}{} & \multicolumn{1}{c}{} & \multicolumn{1}{c}{} & \multicolumn{2}{c}{MARS} & \multicolumn{2}{c}{iLIDS-VID (10 trials)} \\ 
		\cmidrule(r){1-1} \cmidrule(r){2-2} \cmidrule(r){3-3} \cmidrule(r){4-4} \cmidrule(r){5-6} \cmidrule(r){7-8}
		\multicolumn{1}{c}{} & \multicolumn{1}{c}{} & \multicolumn{1}{c}{} & \multicolumn{1}{c}{} & Running Time       & Rank-1      & Running Time          & Rank-1        \\
		1      & $1\mathcal{\times}210$                                                          & 18.05G  &  85.76M  & 4.25h  & 87.33 &  8.00h & 89.87                    \\
		2      & $1\mathcal{\times}210\_105\mathcal{\times}2$                                    & 19.56G  &  93.09M  & 4.45h  & 88.59 &  8.75h & 90.20                    \\
		3      & $1\mathcal{\times}210\_105\mathcal{\times}2\_3\mathcal{\times}70$               & 21.06G  & 100.53M  & 4.70h  & 88.75 &  9.50h & 91.13                    \\
		4      & $1\mathcal{\times}210\_105\mathcal{\times}2\_3\mathcal{\times}70\_6\mathtt{p}$  & 22.59G  & 108.32M  & 5.00h  & 90.22 & 10.70h & 92.07          \\ \hline
	\end{tabular}
\end{table*}

\subsubsection{Performances of Proposed Pyramid with Different Parameters} \label{four-layer-2}
The proposed PiT contains four layers, and each layer adopts different division strategies. The performances of the proposed PiT with different parameters are presented in Table \ref{parameters}. For example, the parameter $1\mathcal{\times}210\_105\mathcal{\times}2\_3\mathcal{\times}70\_6\mathtt{p}$ represents ``no division'' ($1\mathcal{\times}210$), vertical division ($105\mathcal{\times}2$), horizontal division ($3\mathcal{\times}70$), and patch-based division ($6\mathtt{p}$) are employed.

As illustrated in Table \ref{parameters}, the proposed PiT with parameter $1\mathcal{\times}210\_105\mathcal{\times}2\_3\mathcal{\times}70\_6\mathtt{p}$ achieves the best performance. The other two pyramids introduce more fine-grained parts, yet their performances get poor. On the other side, the pyramid with the parameter $1\mathcal{\times}210\_105\mathcal{\times}2\_7\mathcal{\times}30\_14\mathtt{p}$ splits the patch tokens into more horizontal parts than using the parameter $1\mathcal{\times}210\_42\mathcal{\times}5\_3\mathcal{\times}70\_15\mathtt{p}$ and achieves better performance. Although their fourth layers have a close number of parts, more horizontal parts introduce more performance improvement.

\subsubsection{Qualitative Analysis} \label{qualitative}
\begin{figure*}[htp]
	\centering
	\includegraphics[width=1\textwidth]{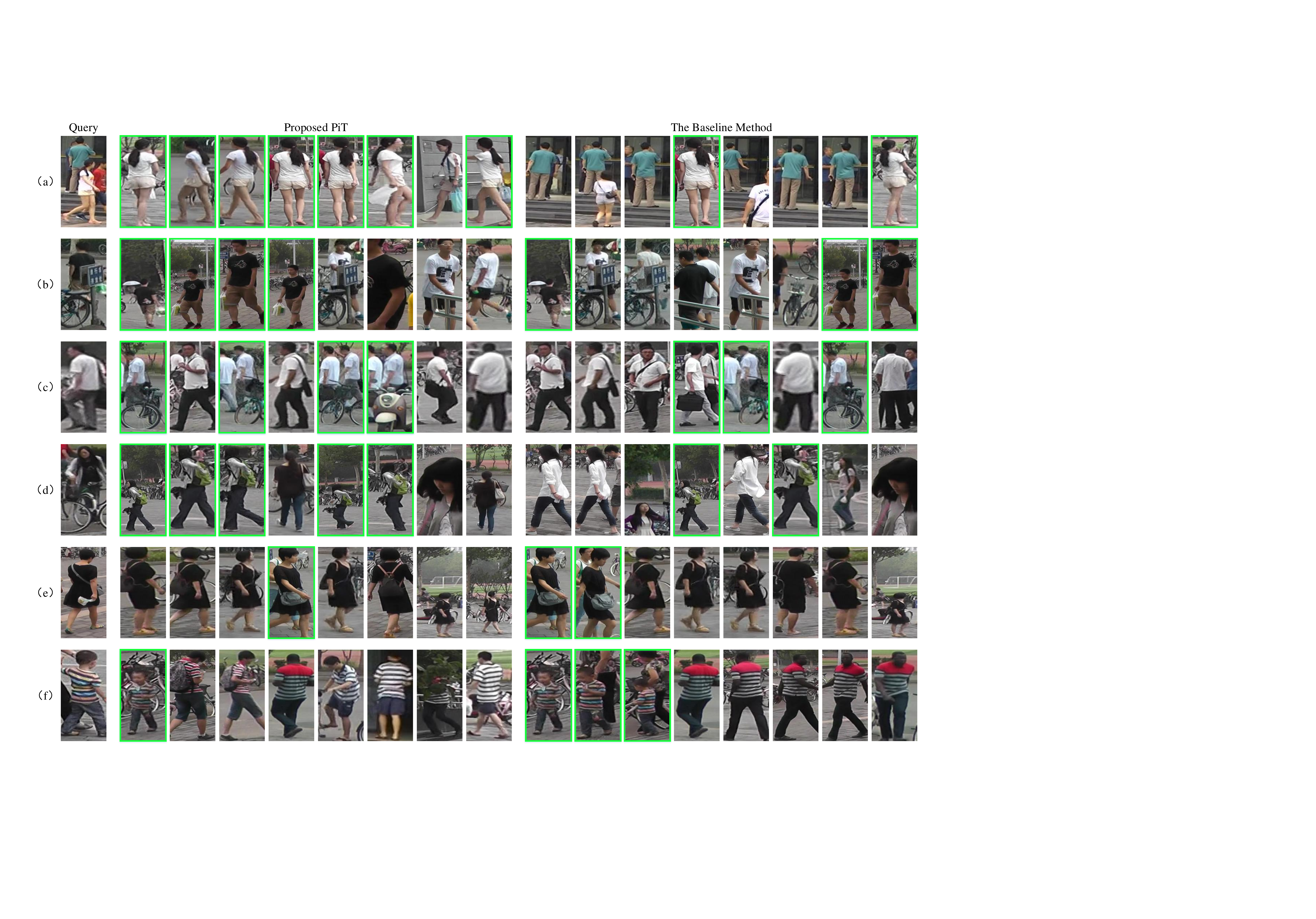}
	\caption{Retrieval examples. Each video is represented by one pedestrian image within it. The query video is selected from MARS benchmark, and the top eight retrieval results for each query are illustrated in this figure. (a)(b)(c)(d) show the successful cases, and (e)(f) show the failed cases. The correct results are in green boxes.}
	\label{fig:retrievalResults}
\end{figure*}

The retrieval examples are illustrated in Fig. \ref{fig:retrievalResults}. One pedestrian image is selected to represent the video for convenience, and the top eight retrieval results for each query are illustrated in this figure. 
We select the query pedestrian video from MARS benchmark according to the Average Precision (AP) value. Fig. \ref{fig:retrievalResults}(a)(b)(c)(d) show the successful cases where the proposed PiT has a better AP than the baseline method, and Fig. \ref{fig:retrievalResults}(e)(f) show the failed cases where the baseline method performs better.

For the successful case in Fig. \ref{fig:retrievalResults}(a)(b), the proposed PiT retrieves many correct videos in the gallery. In contrast, the baseline method retrieves many incorrect results, including a man in a blue shirt in Fig. \ref{fig:retrievalResults}(a) and a road sign in the second and third places in Fig. \ref{fig:retrievalResults}(b). For the failed case in Fig. \ref{fig:retrievalResults}(d), the baseline method puts the correct videos in the top two places. However, the proposed PiT also retrieves pedestrians with similar appearances. 
In Fig. \ref{fig:retrievalResults}, we employ the AP value to determine whether the proposed method is successful or not. For practical application, people usually look for the person of interest from the top-k results, not just the top-1 result. Therefore, the proposed PiT can be utilized effectively for Fig. \ref{fig:retrievalResults}(e)(f).

\begin{figure}[htp]
	\centering
	\includegraphics[width=0.48\textwidth]{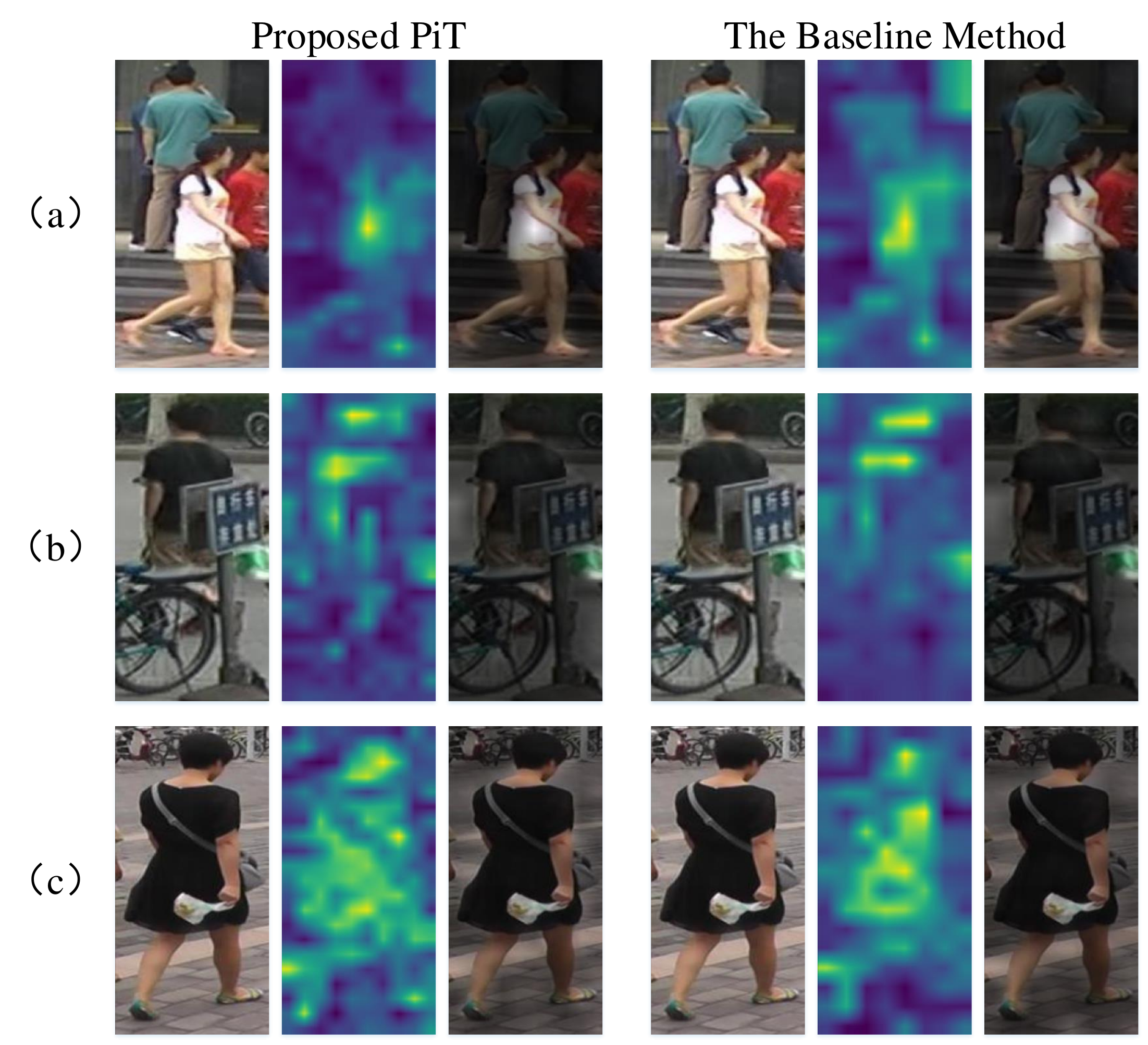}
	\caption{Attention map examples. Three images in each group are the input image, the attention map of this image, and the product result between input image and its attention map.}
	\label{attentions}
\end{figure}

To explore the difference between the proposed PiT and the baseline method, the attention maps of the query pedestrian image are illustrated in Fig. \ref{attentions}. 
With the same input image, the proposed PiT and the baseline method have different attention maps. The baseline method cannot extract the fine-grained local features. Therefore, it cannot reduce the unfavorable impacts from the man in a blue shirt in Fig. \ref{attentions}(a) and the road sign in Fig. \ref{attentions}(b).
Therefore, more fine-grained feature representation can help the model recognize the pedestrian of interest.

\subsubsection{Computation Complexity and Running Time} \label{sec:computation}
Table \ref{computation} adopts Multiply-ACcumulate operations (MACs) and trainable parameters to show the computation complexity. We use one 24G NVIDIA TITAN RTX GPU to conduct the experiments. For the MARS benchmark, the training and testing processes of the proposed PiT take 5.00 hours, including training 8,298 videos from 625 pedestrian IDs and testing another 11,310 videos for evaluation. For the iLIDS-VID benchmark, the experiments include ten trials to ensure statistical stability, and the total running time takes 10.70 hours. Each trial contains training 300 videos from 150 IDs and testing another 300 videos. In Table \ref{computation}, the proposed PiT has acceptable MACs, trainable parameters, and running time, which have the same order of magnitude as the baseline method. The superiority of the proposed method can be seen from the performance improvement in terms of the Rank-1 metric.

\section{Conclusion} \label{conlusion}
This paper proposes a multi-direction and multi-scale Pyramid in Transformer (PiT) for video-based pedestrian retrieval.
The proposed PiT contains four layers, and each layer applies different division strategies on the patch tokens to generate different-direction parts. The class token and the patch tokens in each generated part are fed to the corresponding transformer layer. In this way, the class token perceives the fine-grained, part-informed features. Then multi-direction and multi-scale features are combined to form a feature pyramid for each pedestrian image. The feature pyramids of pedestrian images belonging to the same video are fused to generate the final feature pyramid. Experimental results on two challenging benchmarks, MARS and iLIDS-VID, show the proposed PiT achieves state-of-the-art results. The comprehensive ablation studies demonstrate the superiority of the proposed multi-direction and multi-scale pyramid structure.

\ifCLASSOPTIONcaptionsoff
\newpage
\fi



\normalem
\bibliographystyle{IEEEtran}
\bibliography{BIB_TII-21-4985.R1}

\begin{thebibliography}{10}
\providecommand{\url}[1]{#1}
\csname url@samestyle\endcsname
\providecommand{\newblock}{\relax}
\providecommand{\bibinfo}[2]{#2}
\providecommand{\BIBentrySTDinterwordspacing}{\spaceskip=0pt\relax}
\providecommand{\BIBentryALTinterwordstretchfactor}{4}
\providecommand{\BIBentryALTinterwordspacing}{\spaceskip=\fontdimen2\font plus
\BIBentryALTinterwordstretchfactor\fontdimen3\font minus
  \fontdimen4\font\relax}
\providecommand{\BIBforeignlanguage}[2]{{%
\expandafter\ifx\csname l@#1\endcsname\relax
\typeout{** WARNING: IEEEtran.bst: No hyphenation pattern has been}%
\typeout{** loaded for the language `#1'. Using the pattern for}%
\typeout{** the default language instead.}%
\else
\language=\csname l@#1\endcsname
\fi
#2}}
\providecommand{\BIBdecl}{\relax}
\BIBdecl

\bibitem{RN632}
M.~Ye, Y.~Cheng, X.~Lan, and H.~Zhu, ``Improving night-time pedestrian
  retrieval with distribution alignment and contextual distance,'' \emph{IEEE
  Transactions on Industrial Informatics}, vol.~16, no.~1, pp. 615--624, 2019.

\bibitem{RN636}
J.~García, A.~Gardel, I.~Bravo, and J.~L. Lázaro, ``Multiple view oriented
  matching algorithm for people reidentification,'' \emph{IEEE Transactions on
  Industrial Informatics}, vol.~10, no.~3, pp. 1841--1851, 2014.

\bibitem{RN687}
X.~Zang, G.~Li, W.~Gao, and X.~Shu, ``Learning to disentangle scenes for person
  re-identification,'' \emph{Image and Vision Computing}, vol. 116, p. 104330,
  2021.

\bibitem{RN635}
Z.~Zeng, Z.~Li, D.~Cheng, H.~Zhang, K.~Zhan, and Y.~Yang, ``Two-stream
  multirate recurrent neural network for video-based pedestrian
  reidentification,'' \emph{IEEE Transactions on Industrial Informatics},
  vol.~14, no.~7, pp. 3179--3186, 2017.

\bibitem{RN191}
Y.~Sun, L.~Zheng, Y.~Yang, Q.~Tian, and S.~Wang, ``Beyond part models: Person
  retrieval with refined part pooling (and a strong convolutional baseline),''
  in \emph{Proceedings of the European Conference on Computer Vision (ECCV)},
  2018, Conference Proceedings, pp. 480--496.

\bibitem{RN318}
Y.~Fu, Y.~Wei, Y.~Zhou, H.~Shi, G.~Huang, X.~Wang, Z.~Yao, and T.~Huang,
  ``Horizontal pyramid matching for person re-identification,'' in
  \emph{Proceedings of the AAAI Conference on Artificial Intelligence},
  vol.~33, 2019, Conference Proceedings, pp. 8295--8302.

\bibitem{RN710}
A.~Dosovitskiy, L.~Beyer, A.~Kolesnikov, D.~Weissenborn, X.~Zhai,
  T.~Unterthiner, M.~Dehghani, M.~Minderer, G.~Heigold, and S.~Gelly, ``An
  image is worth 16x16 words: Transformers for image recognition at scale,'' in
  \emph{International Conference on Learning Representations}, 2020, Conference
  Proceedings.

\bibitem{RN158}
L.~Zheng, Z.~Bie, Y.~Sun, J.~Wang, C.~Su, S.~Wang, and Q.~Tian, ``Mars: A video
  benchmark for large-scale person re-identification,'' in \emph{Proceedings of
  the European Conference on Computer Vision (ECCV)}.\hskip 1em plus 0.5em
  minus 0.4em\relax Springer, 2016, Conference Proceedings, pp. 868--884.

\bibitem{RN386}
T.~Wang, S.~Gong, X.~Zhu, and S.~Wang, ``Person re-identification by video
  ranking,'' in \emph{Proceedings of the European Conference on Computer Vision
  (ECCV)}.\hskip 1em plus 0.5em minus 0.4em\relax Springer, 2014, Conference
  Proceedings, pp. 688--703.

\bibitem{RN708}
R.~Zhou, X.~Chang, L.~Shi, Y.-D. Shen, Y.~Yang, and F.~Nie, ``Person
  reidentification via multi-feature fusion with adaptive graph learning,''
  \emph{IEEE Transactions on Neural Networks and Learning Systems}, vol.~31,
  no.~5, pp. 1592--1601, 2019.

\bibitem{RN686}
X.~Zang, G.~Li, W.~Gao, and X.~Shu, ``Exploiting robust unsupervised video
  person re‐identification,'' \emph{IET Image Processing}, 2021.

\bibitem{RN606}
X.~Liu, P.~Zhang, C.~Yu, H.~Lu, and X.~Yang, ``Watching you: Global-guided
  reciprocal learning for video-based person re-identification,'' in
  \emph{Proceedings of the IEEE/CVF Conference on Computer Vision and Pattern
  Recognition}, 2021, Conference Proceedings, pp. 13\,334--13\,343.

\bibitem{RN607}
R.~Hou, H.~Chang, B.~Ma, R.~Huang, and S.~Shan, ``Bicnet-tks: Learning
  efficient spatial-temporal representation for video person
  re-identification,'' in \emph{Proceedings of the IEEE/CVF Conference on
  Computer Vision and Pattern Recognition}, 2021, Conference Proceedings, pp.
  2014--2023.

\bibitem{RN608}
J.~Liu, Z.-J. Zha, W.~Wu, K.~Zheng, and Q.~Sun, ``Spatial-temporal correlation
  and topology learning for person re-identification in videos,'' in
  \emph{Proceedings of the IEEE/CVF Conference on Computer Vision and Pattern
  Recognition}, 2021, Conference Proceedings, pp. 4370--4379.

\bibitem{RN709}
M.~Ye, J.~Shen, G.~Lin, T.~Xiang, L.~Shao, and S.~C. Hoi, ``Deep learning for
  person re-identification: A survey and outlook,'' \emph{IEEE Transactions on
  Pattern Analysis and Machine Intelligence}, 2021.

\bibitem{RN630}
Z.~Liu, J.~Ning, Y.~Cao, Y.~Wei, Z.~Zhang, S.~Lin, and H.~Hu, ``Video swin
  transformer,'' \emph{arXiv preprint arXiv:2106.13230}, 2021.

\bibitem{RN570}
K.~He, X.~Zhang, S.~Ren, and J.~Sun, ``Spatial pyramid pooling in deep
  convolutional networks for visual recognition,'' \emph{IEEE Transactions on
  Pattern Analysis and Machine Intelligence}, pp. 1904--16, 2014.

\bibitem{RN421}
Y.~Zhao, X.~Shen, Z.~Jin, H.~Lu, and X.-s. Hua, ``Attribute-driven feature
  disentangling and temporal aggregation for video person re-identification,''
  in \emph{Proceedings of the IEEE Conference on Computer Vision and Pattern
  Recognition}, 2019, Conference Proceedings, pp. 4913--4922.

\bibitem{RN422}
J.~Li, J.~Wang, Q.~Tian, W.~Gao, and S.~Zhang, ``Global-local temporal
  representations for video person re-identification,'' in \emph{Proceedings of
  the IEEE International Conference on Computer Vision}, 2019, Conference
  Proceedings, pp. 3958--3967.

\bibitem{RN423}
A.~Subramaniam, A.~Nambiar, and A.~Mittal, ``Co-segmentation inspired attention
  networks for video-based person re-identification,'' in \emph{Proceedings of
  the IEEE International Conference on Computer Vision}, 2019, Conference
  Proceedings, pp. 562--572.

\bibitem{RN558}
X.~Jiang, Y.~Gong, X.~Guo, Q.~Yang, F.~Huang, W.-S. Zheng, F.~Zheng, and
  X.~Sun, ``Rethinking temporal fusion for video-based person re-identification
  on semantic and time aspect,'' in \emph{Proceedings of the AAAI Conference on
  Artificial Intelligence}, vol.~34, 2020, Conference Proceedings, pp.
  11\,133--11\,140.

\bibitem{RN616}
Z.~Chen, Z.~Zhou, J.~Huang, P.~Zhang, and B.~Li, ``Frame-guided region-aligned
  representation for video person re-identification,'' in \emph{Proceedings of
  the AAAI Conference on Artificial Intelligence}, vol.~34, 2020, Conference
  Proceedings, pp. 10\,591--10\,598.

\bibitem{RN617}
X.~Li, W.~Zhou, Y.~Zhou, and H.~Li, ``Relation-guided spatial attention and
  temporal refinement for video-based person re-identification,'' in
  \emph{Proceedings of the AAAI Conference on Artificial Intelligence},
  vol.~34, 2020, Conference Proceedings, pp. 11\,434--11\,441.

\bibitem{RN619}
S.~Li, H.~Yu, and H.~Hu, ``Appearance and motion enhancement for video-based
  person re-identification,'' in \emph{Proceedings of the AAAI Conference on
  Artificial Intelligence}, vol.~34, 2020, Conference Proceedings, pp.
  11\,394--11\,401.

\bibitem{RN567}
J.~Liu, Z.~J. Zha, X.~Zhu, and N.~Jiang, ``Co-saliency spatio-temporal
  interaction network for person re-identification in videos,'' in
  \emph{Twenty-Ninth International Joint Conference on Artificial Intelligence
  and Seventeenth Pacific Rim International Conference on Artificial
  Intelligence {IJCAI-PRICAI-20}}, 2020, Conference Proceedings.

\bibitem{RN615}
X.~Zhu, J.~Liu, H.~Wu, M.~Wang, and Z.-J. Zha, ``Asta-net: Adaptive
  spatio-temporal attention network for person re-identification in videos,''
  in \emph{Proceedings of the 28th ACM International Conference on Multimedia},
  2020, Conference Proceedings, pp. 1706--1715.

\bibitem{RN418}
Z.~Zhang, C.~Lan, W.~Zeng, and Z.~Chen, ``Multi-granularity reference-aided
  attentive feature aggregation for video-based person re-identification,'' in
  \emph{Proceedings of the IEEE/CVF Conference on Computer Vision and Pattern
  Recognition}, 2020, Conference Proceedings, pp. 10\,407--10\,416.

\bibitem{RN424}
Y.~Yan, J.~Qin, J.~Chen, L.~Liu, F.~Zhu, Y.~Tai, and L.~Shao, ``Learning
  multi-granular hypergraphs for video-based person re-identification,'' in
  \emph{Proceedings of the IEEE/CVF Conference on Computer Vision and Pattern
  Recognition}, 2020, Conference Proceedings, pp. 2899--2908.

\bibitem{RN425}
J.~Yang, W.-S. Zheng, Q.~Yang, Y.-C. Chen, and Q.~Tian, ``Spatial-temporal
  graph convolutional network for video-based person re-identification,'' in
  \emph{Proceedings of the IEEE/CVF Conference on Computer Vision and Pattern
  Recognition}, 2020, Conference Proceedings, pp. 3289--3299.

\bibitem{RN441}
R.~Hou, B.~Ma, H.~Chang, X.~Gu, S.~Shan, and X.~Chen, ``Vrstc: Occlusion-free
  video person re-identification,'' in \emph{Proceedings of the IEEE Conference
  on Computer Vision and Pattern Recognition}, 2019, Conference Proceedings,
  pp. 7183--7192.

\bibitem{RN614}
R.~Hou, H.~Chang, B.~Ma, S.~Shan, and X.~Chen, ``Temporal complementary
  learning for video person re-identification,'' in \emph{European Conference
  on Computer Vision}.\hskip 1em plus 0.5em minus 0.4em\relax Springer, 2020,
  Conference Proceedings, pp. 388--405.

\bibitem{RN618}
X.~Gu, H.~Chang, B.~Ma, H.~Zhang, and X.~Chen, ``Appearance-preserving 3d
  convolution for video-based person re-identification,'' in \emph{European
  Conference on Computer Vision}.\hskip 1em plus 0.5em minus 0.4em\relax
  Springer, 2020, Conference Proceedings, pp. 228--243.

\bibitem{RN565}
G.~Chen, Y.~Rao, J.~Lu, and J.~Zhou, ``Temporal coherence or temporal motion:
  Which is more critical for video-based person re-identification?'' in
  \emph{European Conference on Computer Vision}.\hskip 1em plus 0.5em minus
  0.4em\relax Springer, 2020, Conference Proceedings, pp. 660--676.

\bibitem{RN620}
X.~Jiang, Y.~Qiao, J.~Yan, Q.~Li, W.~Zheng, and D.~Chen, ``Ssn3d:
  Self-separated network to align parts for 3d convolution in video person
  re-identification,'' in \emph{Proceedings of the AAAI Conference on
  Artificial Intelligence}, vol.~35, 2021, Conference Proceedings, pp.
  1691--1699.

\bibitem{RN711}
A.~Porrello, L.~Bergamini, and S.~Calderara, ``Robust re-identification by
  multiple views knowledge distillation,'' in \emph{European Conference on
  Computer Vision}.\hskip 1em plus 0.5em minus 0.4em\relax Springer, 2020,
  Conference Proceedings, pp. 93--110.

\end{thebibliography}
\end{document}